\documentclass{IOS-Book-Article}
\usepackage{mathptmx}
\usepackage{soul}\setuldepth{article}
\usepackage{subcaption}
\usepackage{amsmath}
\def\hb{\hbox to 11.5 cm{}}

\usepackage{graphicx}
\usepackage{algpseudocode}
\usepackage{algorithm}

\algnewcommand\algorithmicforeach{\textbf{for each}}
\algdef{S}[FOR]{ForEach}[1]{\algorithmicforeach\ #1\ \algorithmicdo}
\algdef{SE}[REPEATN]{RepeatN}{End}[1]{\algorithmicrepeat\ #1 \textbf{times}}{\algorithmicend}

\begin{document}
	
	\pagestyle{headings}
	\def\thepage{}
	\begin{frontmatter}              

		\title{Hierarchical Support Vector State Partitioning for Distilling Black Box Reinforcement Learning Policies}
		
		
		\author[A, B]{\fnms{Senne} \snm{Deproost}}
	\author[A]{\fnms{Mehrdad} \snm{Asadi}}
and
\author[A, B]{\fnms{Ann} \snm{Nowé}}

\runningauthor{S. Deproost et al.}
\address[A]{Vrije Universiteit Brussel}
\address[B]{Flanders Make}

\begin{abstract}
We introduce State Vector Space Partitioning (SVSP), a novel method to mimic a black-box reinforcement learning policy using a set of human-interpretable sub-policies. 
By partitioning a distillation dataset of state–action pairs with linear support vector machine splits, SVSP constructs a compact and structured representation of the original policy where linear models can be interpreted as a measure of feature importance. 
Our method improves mean return by +7.4\% over previous critic-driven state partitioning attempts such as Voronoi State Partitioning (VSP) and +2.8\% over the original TD3 policy, while reducing the number of required sub-policies against VSP by 82.1\%. Our results pave the path towards a more flexible form of distillation where both the decision boundary and surrogate models can be chosen within a margin of the original black box behavior.
\end{abstract}

\end{frontmatter}

\markboth{April 2026\hb}{April 2026\hb}

\section{Introduction}
State-of-the-art Reinforcement Learning (RL) methods built upon deep neural networks, making them hard to interpret, discouraging users from deploying the learned behavior into the production system. Aside this, it is not straightforward to decide upon a universal explanation that suits both experts (such as control engineers) and users (with reduced background knowledge). 
One way to address this issue is to produce locally interpretable models mimicking the original black-box decision function with more explainable linear representation, standing in as a measure of feature importance~\cite{ribeiro2016should}.
Global interpretability requires distilling policies into transparent surrogate decision trees via imitation learning \cite{coppens2019distilling, kohleretal2024InterpretableEditable}.
Recent work has explored the use of RL-specific components, such as value networks and critics, to guide surrogate modeling~\cite{deproost2024human}. In parallel, advances in tree-based methods suggest that complex policies can be approximated by decomposing the state space into regions governed by simple sub-policies~\cite{bastani2018verifiable}.  
In this work, we propose to leverage theoretical insights from Support Vector Machines (SVMs) \cite{blanco2023multiclass} to split the state space into regions where a specialized model performs comparable behavior to the original black-box policy. 
\vspace{-5mm}
\section{Method}
The main idea of Support Vector State Partitioning (SVSP) is the hierarchical splitting of the state space $S$ into regions $r_i \in R$ where a subpolicy $\tilde{\pi_r}$, represented by a human-interpretable class of model, can perform behavior that is close enough to the original RL policy $\pi$. The critic $Q_{\pi}$, which is a value estimator from the original policy that predicts the expected return of an action $a \in A$ in a state $s \in S$, labels the subset in the region considered not sufficiently covered by the subpolicy. The state routing is learned by a binary SVM classifier, deciding whether or not to output an action with the current subpolicy or one deeper in the hierarchy.

\begin{algorithm}
	\footnotesize
	\caption{Support Vector State Partitioning}
	\begin{algorithmic}[1]
		
		\Require State-action pairs $\left\langle S_\pi, A_\pi \right\rangle$, critic network $Q_\pi$
		\State Initialize list of tree nodes with $r_0$ representing an arbitrary subpolicy
		
		\For{$n = 0$ to $\texttt{n\_iteration}$} 
		
		\State Fit subpolicy $\tilde{\pi}_{r}$ on $\left\langle S_{r}, A_{r} \right\rangle$
		\State $A_{\tilde{\pi}_{r}} \gets \tilde{\pi} \left( {S_{r}} \right)$
		\State Evaluate $\left\langle S_{r}, A_{\tilde{\pi}_{r}} \right\rangle$ using $Q_\pi$
		\State Label states $l_i = 1$ where $Q(S_{r}, A_{\tilde{\pi}_{r}}) / V(S_{r}) > \texttt{value\_threshold}$, 0 otherwise
		\State Fit SVM classifier $f_{r}(s) \rightarrow \{ 0, 1\}$ on labeled states
		\State Add child node $c$ with parent $r$
		\State $\left\langle S_{c}, A_{c} \right\rangle \gets \{(\left\langle s_i, a_i \right\rangle, l_i) \in \left\langle S_{r}, A_{r} \right\rangle \mid l_i = 0\}$ 
		\For{$c_i$ in $r.\texttt{children}$}
		\State Repeat
		\EndFor
		\EndFor

	\end{algorithmic}
	\label{alg:vsp}
\end{algorithm}

As shown in Algorithm \ref{alg:vsp}, we start by fitting the chosen class of model on the entire dataset. 
Secondly, the critic evaluates for each state the behavior of the model by predicting the value of the predicted action for that state. Using a ratio $Q(s, a) / V(s)$, with $V$ being the state value, $Q$ the action-value, we compute a relative advantage of the model choosing one action over all possible actions in that state. If this value is above a predetermined threshold
\footnote{The threshold indicates how large the relative difference between predicted action value and state value should be. A threshold of 1 indicates average behavior, while thresholds above 1 enforces higher expectations from the critic to be considered for the subpolicy}, we consider the behavior produced good enough w.r.t. the learned return expectation of the critic (labeled as 1). Those below the threshold are considered too complex for the model to perform on (labeled 0).  After this labeling, the Linear SVM is tasked with finding the decision boundary that best splits the data over those labels. The process is then repeated on the zero-labeled dataset until a predefined max-depth is reached. 
\vspace{-3mm}
\section{Validation}
To evaluate our method, we used the LunarLanderContinuous benchmark from Gymnasium. This problem is an RL task, where an agent must land a spacecraft safely on a designated pad by controlling its thrusters. 
An original TD3 agent from StableBaselines3 \cite{baselines} is trained, and the state-actions of 1000 episodes are recorder. 
Validation is done on 20 independent experiments. 
We compare our results to Voronoi State Partitioning (VSP) \cite{deproostetal2026CriticDrivenVoronoiQuantization}, which is a similar critic-driven partitioning algorithm that uses nearest neighbor lookups to perform the routing to a linear subpolicy.

\begin{table}[h]
	\caption{Performance of the original TD3 agent, the VSP, and the proposed (SVSP) methods. SVSP both outperforms TD3 and VSP with a small set of 10 linear functions.}
	\label{tab:performances}
	\centering
	\resizebox{\textwidth}{!}{
		\begin{tabular}{|c|c|c|c|c|c|}
			\hline
			\textbf{Environment}   & $\operatorname{\textbf{Return}}_{\operatorname{TD3}}$ & $\operatorname{\textbf{Return}}_{\operatorname{VSP}}$      & $\operatorname{\textbf{Return}}_{SVSP}$ & \#\textbf{Sub-Policies} (VSP) & \#\textbf{Sub-Policies} (SVSP)\\
			\hline
			LunarLander   & $161.8 \pm 8.28$& $154.8 \pm 24.91$&  $166.3 \pm 31.07$ & $55.9 \pm 3.71$ & $10$ \\
				\hline     
		\end{tabular}}
	\end{table}
	
	The results of both the VSP and SVSP method are given in Table \ref{tab:performances}. Our method outperforms VSP whilst using only a fifth of the linear subpolicies. Compared to VSP, a larger variance in returns was observed. Aside from this, we notice an improvement compared to the original TD3 policy.
	
	\begin{figure}[]
		\centering
		
		\begin{subfigure}[t]{0.48\textwidth}
			\centering
			\includegraphics[width=\linewidth]{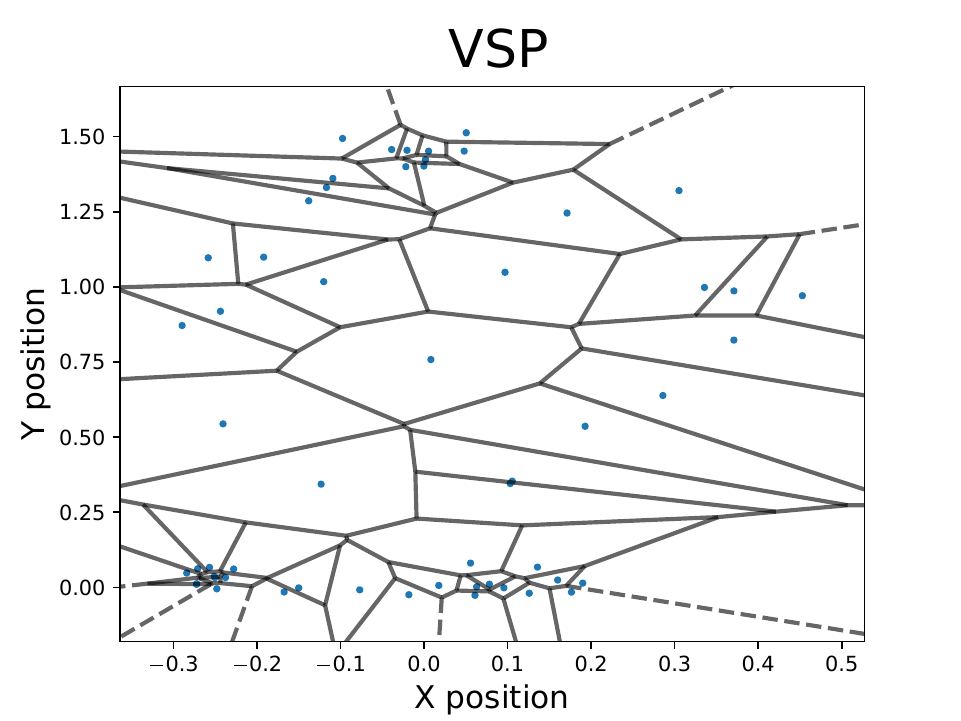}
			\label{fig:vsp}
		\end{subfigure}
		\hfill
		\begin{subfigure}[t]{0.48\textwidth}
			\centering
			\includegraphics[width=\linewidth]{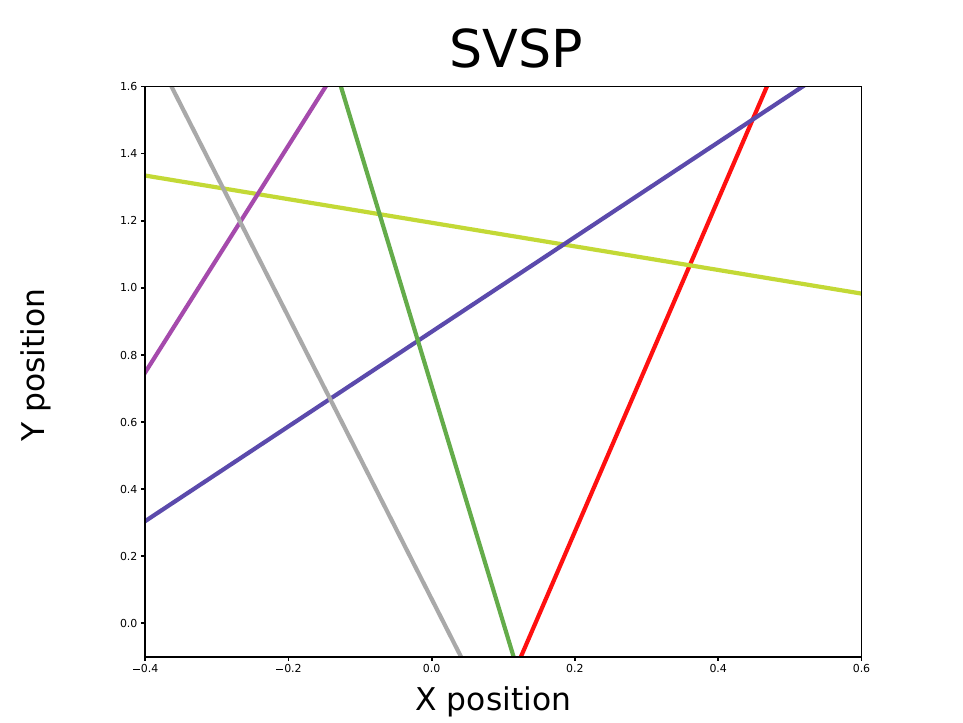}
			\label{fig:svsp}
		\end{subfigure}
		
		\caption{Comparison of the decision boundaries between VSP and SVSP for Lunar Lander. Both the partitioning on the X and Y positions is given with 0 for all other state variables. We notice that SVSP has decision boundaries that are significantly reduced in complexity.}
		\label{fig:comparison}
	\end{figure}
	
	\vspace{-3mm}
	\section{Conclusion and Future Work}
	We introduce State Vector Space Partitioning (SVSP), a novel method to mimic a black box Reinforcement Learning policy with a set of human-interpretable subpolicies. By splitting a distillation dataset of state-actions from the original policy, we show that we partition the space using support vector machines with linear splits.  Our validation on a well-known control benchmark demonstrates that SVSP achieves both higher returns and significantly lower policy complexity. With performance gains of 2.8–7.4\% and an 82.1\% reduction in sub-policies, it captures complex behaviors more efficiently than baseline approaches. These findings highlight SVSP as a practical and interpretable alternative for policy distillation. 
	Future work would also include a user study where the interpretability of different classes of subpolicies can be empirically evaluated during creation and usage.
	
	\section*{Acknowledgment}
	This research received funding from the Flanders Research Foundation via FWO S007723N (CTRLxAI) and FWO G062819N (Explainable Reinforcement Learning). M. A. and A. N. were supported by the Flemish AI Research Program and PEER project (EU Horizon Grant 101120406).


\begin{thebibliography}{99}
	
		\bibitem{ribeiro2016should}Ribeiro, M., Singh, S. \& Guestrin, C. " Why should i trust you?" Explaining the predictions of any classifier. {\em Proceedings Of The 22nd ACM SIGKDD International Conference On Knowledge Discovery And Data Mining}. pp. 1135-1144 (2016)
		
		\bibitem{2025explainable}Deproost, S., Steckelmacher, D. \& Nowé, A. Explainable RL Policies by Distilling to Locally-Specialized Linear Policies with Voronoi State Partitioning. {\em ArXiv Preprint ArXiv:2511.13322}. (2025)
		
		
		\bibitem{deproostetal2026CriticDrivenVoronoiQuantization}Deproost, S., Steckelmacher, D. \& Nowé, A. Critic-Driven Voronoi-Quantization for Distilling Deep RL Policies to Explainable Models.  (2026,5,14), http://arxiv.org/abs/2605.14897
		
		
		\bibitem{kohleretal2024InterpretableEditable}Kohler, H., Delfosse, Q., Akrour, R., Kersting, K. \& Preux, P. Interpretable and Editable Programmatic Tree Policies for Reinforcement Learning.  (2024,10,28)
		\bibitem{coppens2019distilling}Coppens, Y., Efthymiadis, K., Lenaerts, T., Nowé, A., Miller, T., Weber, R. \& Magazzeni, D. Distilling deep reinforcement learning policies in soft decision trees. {\em Proceedings Of The IJCAI 2019 Workshop On Explainable Artificial Intelligence}. pp. 1-6 (2019)
		\bibitem{blanco2023multiclass}Blanco, V., Japón, A. \& Puerto, J. Multiclass optimal classification trees with svm-splits. {\em Machine Learning}. \textbf{112}, 4905-4928 (2023)
		\bibitem{baselines}Dhariwal, P., Hesse, C., Klimov, O., Nichol, A., Plappert, M., Radford, A., Schulman, J., Sidor, S., Wu, Y. \& Zhokhov, P. OpenAI Baselines. {\em GitHub Repository}. (2017), https://github.com/openai/baselines
		\bibitem{bastani2018verifiable}Bastani, O., Pu, Y. \& Solar-Lezama, A. Verifiable reinforcement learning via policy extraction. {\em Advances In Neural Information Processing Systems}. \textbf{31} (2018)
		\bibitem{deproost2024human}Deproost, S., Steckelmacher, D. \& Nowé, A. Human-Readable Programs as Actors of Reinforcement Learning Agents Using Critic-Moderated Evolution. {\em ArXiv Preprint ArXiv:2410.21940}. (2024)
		
	
	\end{thebibliography}
\end{document}